\documentclass[letterpaper,10pt,conference]{ieeeconf}

\IEEEoverridecommandlockouts
\let\labelindent\relax
\usepackage{enumitem}
\usepackage{amssymb}
\usepackage{xcolor}
\usepackage{booktabs}
\usepackage{multirow}
\usepackage{float}
\usepackage{lipsum}
\usepackage{graphicx}
\usepackage{amsmath}
\usepackage[mathcal]{eucal}
\usepackage{caption}
\usepackage{stfloats}
\usepackage{cuted}
\usepackage{cite}
\usepackage{xspace}
\usepackage{indentfirst}
\usepackage{arydshln}
\usepackage[pdftex,hidelinks]{hyperref}
\hypersetup{
	colorlinks=false,
	pdfborder={0 0 0},
	breaklinks=true
}
\usepackage{orcidlink}

\usepackage{fancyhdr}

\newcommand{\acceptedpreprintnotice}{%
	\textcolor{red}{\small
		Accepted preprint. This manuscript has been accepted for presentation at the 35th IEEE International Conference on Robot and Human Interactive Communication (RO-MAN 2026). Acceptance notification date: 30 May 2026. The final published version is pending.}%
}

\makeatletter

\let\ACorigbibitem\bibitem
\def\bibitem{\@ifnextchar[\AClbibitem\ACbibitem}

\def\ACbibitem#1{%
	\ACorigbibitem{#1}%
	\hypertarget{bib:#1}{}%
}

\def\AClbibitem[#1]#2{%
	\ACorigbibitem[#1]{#2}%
	\hypertarget{bib:#2}{}%
}

\makeatother

\newcommand{\linkcite}[1]{\unskip\hyperlink{bib:#1}{\cite{#1}}\xspace}

\newcommand{\textcitelink}[2]{\hyperlink{bib:#1}{#2}}

\definecolor{boldcolor}{gray}{0.2} 
\newcommand{\lightbold}[1]{\textbf{\textcolor{boldcolor}{#1}}}

\begin{document}

\title{\LARGE \bf
Multimodal Voice Activity Projection for Turn-Taking in Social Robots with Voice-Activity-Related Pretrained Encoders
}

\author{Antonio Cano$^{1,2}$\,\orcidlink{0000-0002-0435-4987}\label{author:AC},
	Guillermo Pérez$^{1}$\,\orcidlink{0000-0002-8358-996X}\label{author:GP},
	Luis Merino$^{2}$\,\orcidlink{0000-0003-4927-8647}\label{author:LM} and
	Randy Gomez$^{3}$\,\orcidlink{0000-0002-3191-6818}\label{author:RG}
	\thanks{$^{1}$ Antonio Cano and Guillermo Perez are with 4i Intelligent Insights, Seville, Spain.
    {\tt\small \href{mailto:a.cano@4i.ai}{a.cano@4i.ai},
    \href{mailto:g.perez@4i.ai}{g.perez@4i.ai}}}%
    \thanks{$^{3}$ Antonio Cano is with Universidad de Sevilla, Seville, Spain.
    {\tt\small \href{mailto:acano4@us.es}{acano4@us.es}}}%
    \thanks{$^{2}$ Luis Merino is with Universidad Pablo de Olavide, Seville, Spain.
    {\tt\small \href{mailto:lmercab@upo.es}{lmercab@upo.es}}}%
    \thanks{$^{3}$ Randy Gomez is with Honda Research Institute Japan, Saitama, Japan.
    {\tt\small \href{mailto:r.gomez@jp.honda-ri.com}{r.gomez@jp.honda-ri.com}}}%
}

\maketitle

\pagestyle{fancy}
\fancyhf{}
\fancyhead{}
\fancyfoot[C]{\acceptedpreprintnotice}
\renewcommand{\headrulewidth}{0pt}
\renewcommand{\footrulewidth}{0pt}
\thispagestyle{fancy}

\begin{abstract}

Turn-taking prediction is a key requirement for social robots involved in human-human interaction, particularly in mediator settings, where the robot must anticipate conversational dynamics rather than merely react to pauses. This work presents a Multimodal Voice Activity Projection (MM-VAP) framework that extends the original audio-only VAP formulation to synchronized audio-visual inputs while preserving its self-supervised future-projection objective. The proposed approach builds on pretrained audio-visual backbones originally optimized for speech-related tasks and adapts them through Low-Rank Adaptation to the multimodal turn-taking problem. After independent speaker encoding, an inter-speaker attention stage models the relational dynamics required to project future voice activity. In addition, a semantic consistency loss is introduced to regularize the 256-state output space according to higher-level dialogue activity patterns. Experiments on NoXi and NoXi+J showed improvements over the current baselines, particularly for some turn-taking events. Additional evaluation on the Haru EDR corpus further supported the suitability of this direction for mediation-oriented human-robot interaction.
\vspace{-0.5mm}

\noindent The source codes and
pretrained models are available at \textcolor{gray}{\href{https://github.com/acano15/MM-VAP}{https://github.com/acano15/MM-VAP}}.
\end{abstract}
\section{Introduction}
\vspace{-1.5mm}
Turn-taking is the universal mechanism of human communication for structuring spoken interaction and coordinating speaker and listener roles \linkcite{sacksSimplestSystematicsOrganization1974}. To achieve effective Human–Robot Interaction (HRI), conversational robots must account for this human-like coordination across verbal and non-verbal behavior. In social robotics, interaction is inherently turn-based \linkcite{palinkoSocialRobotics16th2025}, so accurate interpretation of conversational flow and timing is necessary to enable fluid and effective communication.

\begin{figure}[tb!]
	\centering
	\includegraphics[width=0.75\columnwidth]{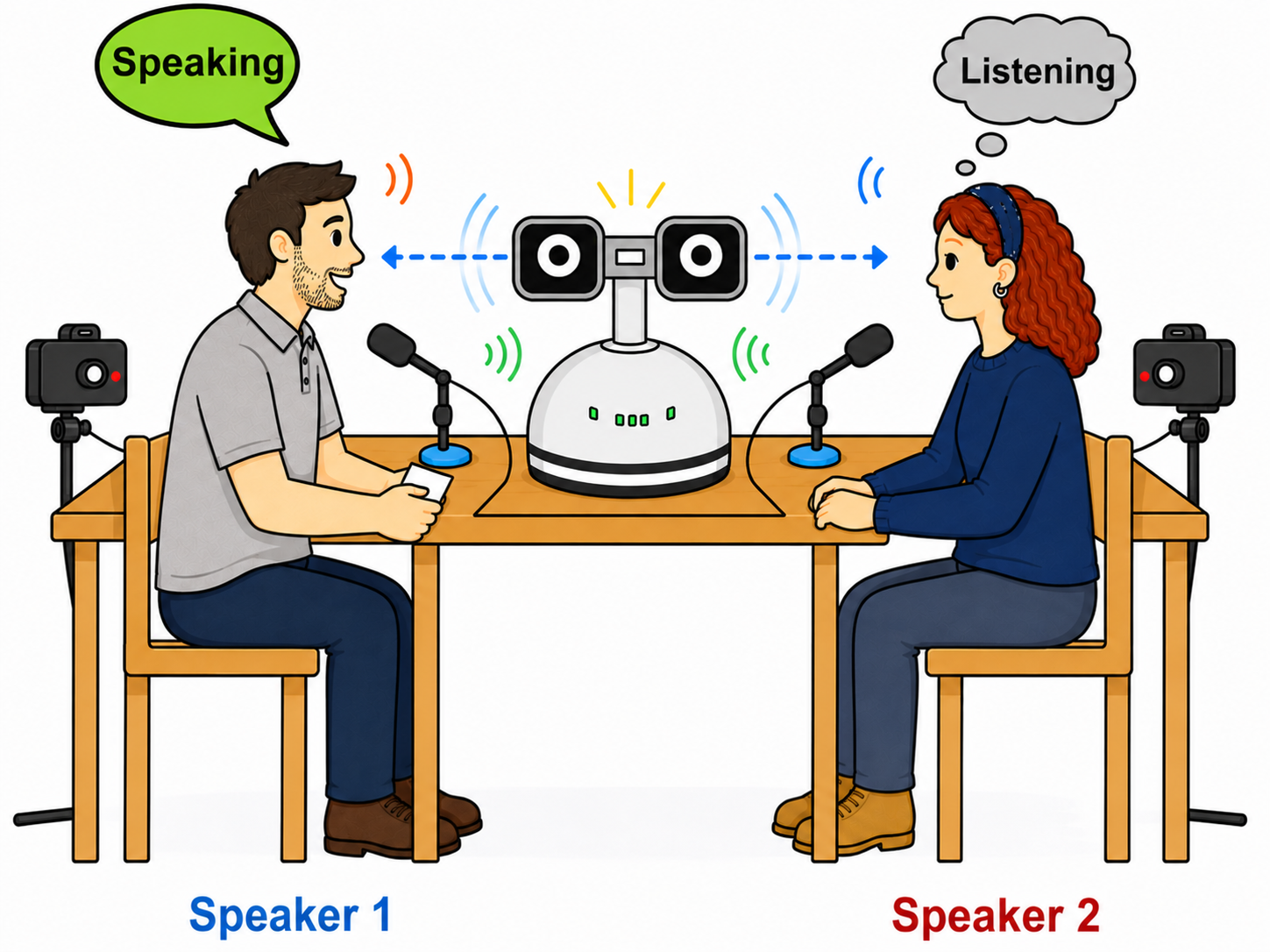}
	\caption{Haru operating as social mediator in two-party interaction setting \protect\linkcite{cruzWhenHowExpress2025}.}
	\label{fig:scenario}
	\vspace*{-7.mm}
\end{figure}

This work extends previous research \linkcite{canoImprovingTurntakingSocialb} within the Haru social robot project, specifically when Haru acts as an embodied social mediator that supports human-human interactions \linkcite{cooperDesignSocialFeatures2024,levinsonHaruCareNetwork2025,yiBuildingFriendshipsBorders2025}. In this scenario, Haru is not conceived as one of the main interlocutors, but instead as a background social mediator (Fig. \ref{fig:scenario}) that helps regulate conversational flow, balance participation, manage silences, and support socio-emotional behaviors such as active listening, engagement, and empathy. This mediator role motivates the present research toward enabling Haru to better understand and manage the ongoing interaction, with particular emphasis on how the conversational floor evolves over time and anticipate when it may change.

Nevertheless, current dialogue systems are not yet sufficiently accurate at managing turn-taking, and this remains an active area of research because users frequently experience the effects of such failures. Typical problems include interruptions or overlaps due to speaking before or after a turn has ended, unnaturally long delays caused by late responses, and incorrect answers resulting from misunderstanding the user’s input. Furthermore, because human communication is inherently multimodal \linkcite{baeMultimodalTransformerModels2025}, these limitations, together with the lack of multimodal perception skills, may cause the robot to gaze at the wrong person or respond to irrelevant background speech.

Addressing this issue requires a structured view of how turns are formed, maintained, and exchanged during interaction. In conversation analysis, spoken interaction is described in terms of Turn-Constructional Units (TCUs), whose possible completion defines a Transition-Relevant Place (TRP) that may lead to a turn shift, a turn hold, or floor closure. Most industrial systems remain \textit{reactive} \linkcite{russellVisualCuesEnhance2025}, using heuristic silence-based rules to detect the End-of-Turn (EoT), normally around 700 ms \linkcite{skantzeTurntakingConversationalSystems2021}. Studies in dialogue systems and dialogue robots \linkcite{meenaDatadrivenModelsTiming2014,boschTemporalAspectsTurn2005,patamiaTurnTakingModellingConversational2025,skantzeTurntakingConversationalSystems2021} have shown that such approaches are insufficient for modeling conversational dynamics because speakers may produce longer pauses within their own turn, known as Inter-Pausal Units (IPUs) \linkcite{heldnerPausesGapsOverlaps2010}.
%
%
Human turn-taking in conversations follows a different principle than current Spoken Dialogue Systems (SDSs). A key distinction is that humans do not simply wait for explicit cues indicating that the interlocutor is ready to yield the floor; instead, they continuously predict ongoing events and adapt their role as listener or speaker accordingly. This genuinely predictive behavior is supported by the temporal constraints of speech generation: producing speech typically requires around 600 ms to 1500 ms \linkcite{levinsonTimingTurntakingIts2015, batesTimedPictureNaming2003}, and the time between turns, as well known as Floor Transfer Offset (FTO), usually occurs around 200 ms of silence (in some cultures may approach 0 ms \linkcite{leeMultimodalTurnAnalysis2023}). This temporal mismatch indicates that turn-taking depends on prediction, with multimodal cues guiding anticipation of TRPs and coordinating responses with minimal delay.

Inspired by this behavior, Predictive Turn-Taking Models (PTTMs) have been proposed as neural models that continuously project future conversational activity at frame level, mostly in two-party spoken interaction \linkcite{skantzeTurntakingConversationalSystems2021}. Among current approaches, Voice Activity Projection (VAP) \linkcite{ekstedtVoiceActivityProjection2022} is positioned as one of the most effective formulations for continuous turn-taking prediction, including real-time implementations. VAP projects frame-by-frame the voice activity (VA) of two speakers and models their interaction dynamics to predict turn-taking events. More recently, multimodal extensions of VAP have become an active and promising research direction \linkcite{onishiMultimodalVoiceActivity2025, russellVisualCuesEnhance2025, sagaVoiceActivityProjection2025, linPredictingTurnTakingBackchannel2025}, where the incorporation of non-verbal information offers further opportunities for improvement, despite the challenges that still remain in practical spoken dialogue system environments.

To contribute to this research line, this work proposes a transformers-based MultiModal Voice Activity Projection (MM-VAP) model built on specialized pretrained encoders. The main contributions are the following:
\vspace{-2.5mm}
\begin{itemize}[leftmargin=*]
	\item An audio-visual \textbf{MM-VAP} architecture that integrates \textbf{VA-related pretrained backbones} while preserving their native representational and fusion capabilities for improved VA projection.
	
	\item A parameter-efficient adaptation strategy based on \textbf{Low-Rank Adaptation} for specializing the encoders to the target turn-taking task.
	
	\item \textbf{Improved performance} on mediation-relevant turn-taking states, highlighting the benefits of the proposed pipeline.
	
	\item The \textbf{first} application of MM-VAP in a robotics context, with an \textbf{empirical} evaluation of the proposed approach for predictive turn-taking in the Haru social mediator setting.
	
\end{itemize}
\section{Related Work} \label{related}


Turn-taking prediction has been extensively studied in SDS, HRI, and, more recently, in VAP-based approaches. This section reviews the most relevant prior work and the key considerations that motivated the presented approach.

Recent multimodal speech and language research \linkcite{maScenesMechanisticInterpretability2026,chenGRPOGuidedModalitySelection,songLoRAWhisperParameterEfficientExtensible2024,wuM3SDMultimodalMultiscenario2025,muMultimodalLargeLanguage2025, palaskarMultimodalLargeLanguage2024, meiRobustAdaptationLarge} has shown that large pretrained backbones can be efficiently adapted with Low-Rank Adaptation (LoRA), allowing transfer of knowledge to more specific tasks by updating only a small subset of parameters instead of the entire model.   At the same time, multimodal specialization still depends on attention mechanisms to model cross-modal dependencies. Based on this, special attention is given to two backbones: TalkNet \linkcite{taoSomeoneSpeakingExploring2021}, which uses cross-attention and self-attention for Active Speaker Detection (ASD), and WhisperFlamingo \linkcite{rouditchenkoWhisperFlamingoIntegratingVisual2024}, which combines pretrained Whisper and AV-HuBERT with gated cross-attention for multilingual Audio Visual Speech Recognition (AVSR).

Within the VAP research line, \linkcite{ekstedtVoiceActivityProjection2022} introduced VAP as a self-supervised prediction problem that outputs the VA of two speakers. The original model combines an encoder based on a pretrained CPC backbone operating on the raw audio waveform with a second encoder that incorporates the current VA frame and its history, followed by a transformer that projects future VA. This formulation has remained a central reference for turn-taking modeling in SDSs and most subsequent studies have preserved its core idea with only minor modifications. For example, \linkcite{elmersTriadicMultipartyVoice2025} extended VAP to three-party interaction by reducing the projection space to two bins per speaker, \linkcite{inoueYeahOhContinuous} adapted it to binary and three-class backchannel (BC) prediction and \linkcite{ishiiPredictingEndofturnBackchannel2025} modified the loss function by incorporating BC and EoT terms to balance multitask objectives. Regarding multilinguality, \linkcite{inoueMultilingualTurntakingPrediction2024} showed that monolingual models generalize poorly to unseen languages in contrast with training on multiple languages, whose performance was comparable to the original language-specific implementations. Complementing this finding, \linkcite{satoInvestigatingLanguageIndependence2024} argued that those cross-lingual differences are better explained by dataset discrepancies, particularly inconsistencies in speech segmentation labels, than language itself. Beyond offline benchmarks, VAP has also shown practical value, as its continuous frame-level formulation and lightweight inference make it suitable for real-time (RT) implementation. For instance, \linkcite{inoueRealtimeContinuousTurntaking2024} demonstrated the feasibility of continuous turn-taking prediction by shortening the transformer input (1s), with only limited performance degradation. However, according to \linkcite{patamiaTurnTakingModellingConversational2025}, its effectiveness still depends on sensor reliability and synchronization quality. 


In the field of robotics, turn-taking has shown promising improvements over conventional strategies. The main HRI reference remains \linkcite{skantzeApplyingGeneralTurnTaking2025}, in which one of the speakers is replaced by the robot's own voice so that VAP could provide RT turn-taking event predictions for floor management, interruptions and BC. More recently, \linkcite{inoueNoiseRobustTurnTakingSystem2025} reformulated the training with noisy data to improve robustness and latency in RT environment and validated it on a dialogue robot deployed in a shopping mall.

Concerning multimodal VAP, only a few studies have been reported so far, all of them extending the original acoustic framework to additional modalities under formulations close to the original VAP design. The first multimodal VAP solution was presented in \linkcite{onishiMultimodalVoiceActivity2023} with the addition of a non-verbal branch based on gaze, action units, head pose, and articular-point features. Their results showed that multimodality improved performance across predictive events. This work was later extended in \linkcite{onishiMultimodalVoiceActivity2025} through a broader experimental study, the addition of an overlap prediction metric and the evaluation across multiple corpora. An analogous feature-based direction was followed in \linkcite{russellVisualCuesEnhance2025}, with facial feature positioning as the strongest visual contributor, resulting in improved performance over audio-only baselines. Nevertheless, both approaches relied on engineered descriptors rather than model-based feature extraction. 
Moving toward multimodal backbone-oriented solutions, the unique approach identified was \linkcite{sagaVoiceActivityProjection2025}, in which the authors replaced the hand-crafted facial descriptors with pretrained audio (original CPC) and a facial expression encoder (DEFER). However, the selected encoders remained generic and their pretraining was not related to voice activity, which reduced their potential for VAP. In addition, because the approach used two separate encoders, the inter-modal relation had to be learned from scratch.

Motivated by these limitations, the present work investigates whether multimodal VAP can benefit from specialized speech-activity backbones integrated into its inter-modal attention mechanism to obtain richer, more task-relevant multimodal representations.

\section{Proposed Method}
\begin{figure*}[tbh!]
	\centering
	\noindent
	\includegraphics[width=\textwidth]{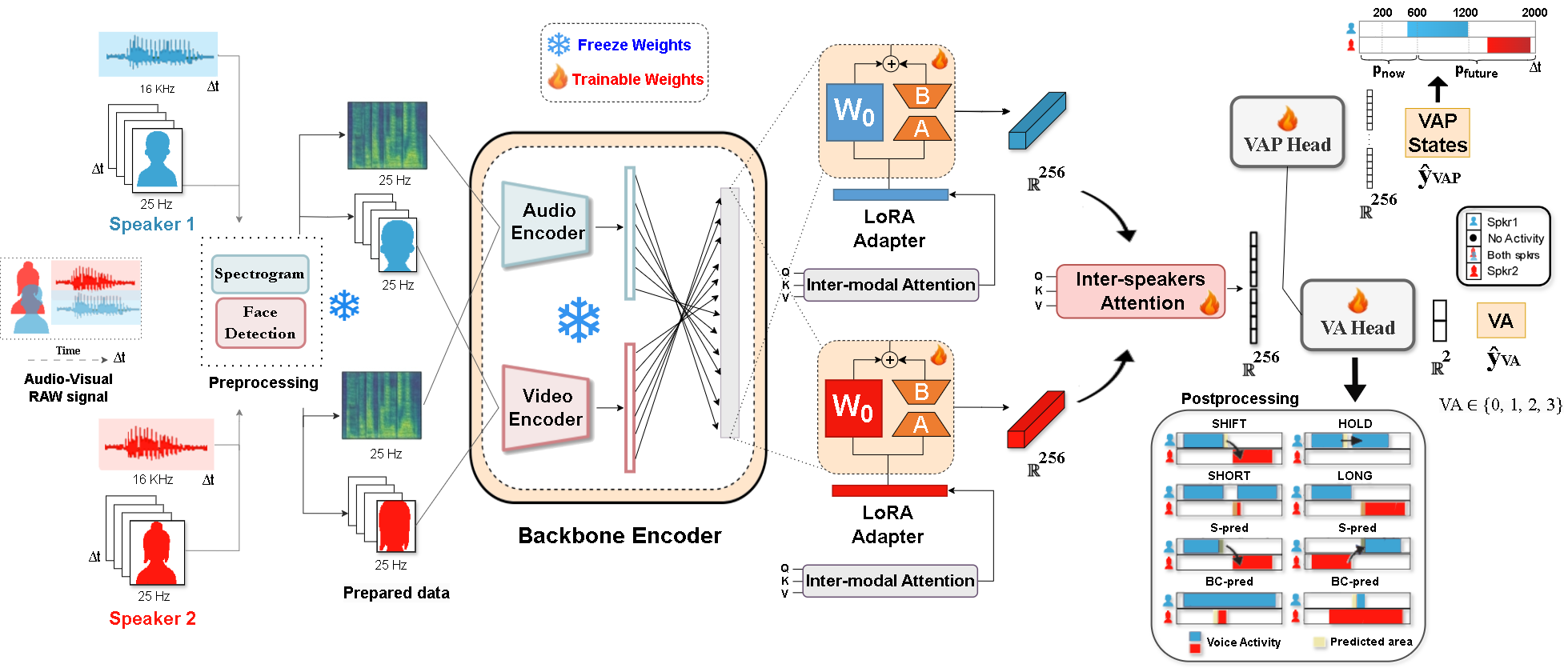}
	\caption{Proposed MM-VAP architecture for dyadic interaction overview. Blue and red paths indicate the flow from raw inputs to latent representations for Speaker 1 and Speaker 2. Snowflake and flame symbols denote frozen and trainable parameters. The bottom-right panel shows how the target turn-taking events are derived from voice activity patterns within $\mathbf{w_{t}^{\Delta t}}$.}
	\label{fig:mmvap_pipelilne}
	\vspace*{-5.85mm}
\end{figure*}

This section describes the proposed multimodal VAP (MM-VAP) framework and its complete pipeline, from synchronized audio-visual input sequences to VA and VAP state projection estimation and final inference of turn-taking events.

\vspace{-1mm}
\subsection{Multimodal Voice Activity Projection (MM-VAP)}
VAP was formally defined \linkcite{ekstedtVoiceActivityProjection2022} as the task of predicting (projecting) the future VA binary state (active/inactive) of each interlocutor in a dialogue. MM-VAP extends the standard definition by incorporating data from multiple modalities, in this case, audio and visual signals. The projection is formulated over a finite temporal horizon $\Delta t$ that spans from the current time step $t$ to a future point $t$ + $\Delta t$. Within this interval, the future binary VA of both speakers is considered jointly, so that future VA is modeled as a single \textit{projection window}, $\mathbf{w_{t}^{\Delta t}}$, that represents the future trajectory of the dialogue and the joint VA of both speakers as a whole. The choice of the window length involves a balance because it should be long enough to be able to capture relevant turn-taking information but no longer than it attempts to represent excessively complex or unnatural future interaction patterns. As in the original implementation, this window is divided into $4$ heterogeneous bins per speaker. The first two bins are shorter, representing the immediate present ($now$), or near future, and the last two bins correspond to longer bins that cover more distant $future$. Since the scenario involves two speakers, the complete projection window is represented by a joint space of $2^{2 N} = 256$ states, in which each state compactly covers every possible configuration of the horizon interaction between the two interlocutors. 

\vspace{-1mm}
\subsection{Proposed MM-VAP model architecture}

As shown in Fig. \ref{fig:mmvap_pipelilne}, the model is organized in five stages: preprocessing, backbone encoding, inter-speaker attention, prediction heads with postprocessing.

First, the raw audio-visual streams are segmented in the $w_{t}^{\Delta t}$ centered at each frame step $t$ with $\Delta t$ horizon separated into speaker-specific signals. For each speaker $i$, the audio waveform is transformed into a spectral representation, and the visual stream is processed by the face detection module that extracts the temporally ordered sequence of cropped face images. Then, both modalities are synchronized to the same temporal resolution and fed together into the pretrained, fine-tuned encoder. Inside the encoder, each modality is first processed by its own corresponding modality encoder branch, yielding speaker representations in their respective latent spaces. The resulting audio and visual embeddings are fused with the backbone's original inter-modal attention mechanism, producing an embedding per speaker: \newline
 $\mathbf{H^i} = \text{Attn}(E_a(X_i^a), E_v(X_i^v))$, where  $H^i \in \mathbb{R}^{256}$ and $\{ Q_i^a = E_a(X_i^a), \quad K_i^v = V_i^v = E_v(X_i^v) \}$.

Once each $\mathbf{H}$ is obtained, the model applies a second multihead cross-attention mechanism between both speakers, referred to as inter-speaker attention, to learn how the current multimodal state of one participant influences the other. $\mathbf{z} = \Phi_{\text{inter-spkrs}} \left( \mathbf{H}^{(1)}, \mathbf{H}^{(2)} \right)$. This stage allows the model to encode turn-taking as a relational process, so the prediction is conditioned on both the speaker’s current state and the other speaker's state. The joint representation $\mathbf{z}$ is then fed to two heads: the first head estimates the multiclass VAP state distribution, and the second head predicts the binary VA probabilities.
Finally, these outputs are postprocessed to compute the target turn-taking event categories from the predicted VAP state distribution.

\vspace{-1mm}
\subsection{Backbone encoders and LoRA adaptation}
The current MM-VAP solution is based on the premise that modeling turn-taking is equivalent to predict the evolution of the speaker's VA dynamics from audio-visual cues. This work hypothesized that a pretrained encoder whose original task is closely related to VA would be more suitable as a starting point for MM-VAP than a pretrained generic multimodal encoder. Under this view, the multimodal encoder should provide latent representations that remain strongly aligned with speech activity and are able to preserve the temporal coordination between audio-visual modalities. 

In addition, the backbone selection was based on two criteria: first, the backbone should achieve high accuracy on its original task related to speech activity, and second, it should include a robust internal attention mechanism between audio-visual modalities. The architecture is modularly designed, so any transformer-based backbone that satisfies these conditions could be integrated. In this work, TalkNet and WhisperFlamingo were selected because they both meet both requirements and provide well proved pretrained multimodal encoders for ASD and AVSR respectively, which are based on speech activity. 

Nevertheless, their original pretrained weights, $W_{av}^{0} \in \mathbb{R}^{dxk}$, remain specialized for the original task, so an adaptation mechanism is required to redirect their representations to the target setting. Given the size and the high variance of the original backbones weights, fully fine-tuning the pretrained model is computationally very inefficient, which motivates the use of Low-Rank Adaptation (LoRA) training strategy. LoRA keeps the original model weights frozen and adds trainable low-rank matrices to selected layers to adapt the backbone to the MM-VAP objective. This strategy is computationally more efficient because it approximates the full objective weights ($\Delta W$) by applying the low-rank decomposition of the $\Delta W$ into a scalar product of two smaller matrices $B \in \mathbb{R}^{dxr}$ and $A \in \mathbb{R}^{rxk}$. Formally, LoRA defines the adapted mapping as:
\begin{equation}
	W_{Av} = W_{Av}^0 + \frac{\alpha}{r} BA, \quad r \ll \min(d, k)  ,
\end{equation}
where $A$ and $B$ are trainable low-rank matrices, $r$ is the adaptation rank and $\alpha$ is the scaling factor. In this context, $r$ controls the flexibility of the adaptation to adapt to the VAP objective and $\alpha$ determines how strongly that adaptation influences the training. 

Within this framework, LoRA enables the specialization of the pretrained backbone to the MM-VAP objective by building on the capabilities acquired during its original training phase and redirecting them toward future VA projection.

\vspace{-1mm}
\subsection{Zero-shot turn-taking events inference}

This work followed the original VAP Self-Supervised Learning (SSL) approach \linkcite{ekstedtVoiceActivityProjection2022}, in which the training is performed without manual annotations of turn-taking events. Specifically, at each frame, the model predicts a probability distribution over the 256 states, and the turn-taking events are obtained in a zero-shot manner by summing the probabilities of the states linked to each conversational outcome. This interpretability is enabled by collapsing the fine-grained state space into a simpler bin-level representation that preserves the same VAP state distribution, with four possible cases: speaker 1 active, speaker 2 active, both active, or neither active. Based on this four-way representation, the aggregated probabilities ${p}_{now}$ and ${p}_{future}$ summarize speaker activity over the near and later regions of the projection window and provide the basis for defining each turn-taking event as follows:
\vspace{-1.9mm}

\begin{itemize}[leftmargin=*]
	\item \textbf{Shift vs Holds (S/H)}: measures who are expected to take the floor after a mutual silence. If the same speaker continues, the event is Hold; if the floor passes to the other speaker, it is Shift.
	
	\item \textbf{Long vs Shorts (L/S)}: describes the duration of the utterance after a valid speaker change. Long means that the new speaker keeps speaking for a sustained period, whereas Short corresponds to a brief response (Backchannel).
	
	\item \textbf{Shift prediction (S-pred)}: checks whether the other speaker is likely to take the floor soon, even though the current speaker is still holding it.
	
	\item \textbf{Backchannel prediction (BC-pred)}: measures whether the non-active (listener) is likely to produce a short response, without the current speaker releasing the turn.
\end{itemize}

\vspace{-2mm}
\subsection{Loss function}
In natural dialogue, many individual VAP state configurations are extremely rare (e.g.: alternating speech in unusual patterns) and the resulting distribution is therefore inherently unbalanced. Moreover, because multiple microstates correspond to the same speaker-activity condition, errors between semantically equivalent microstate predictions do not usually affect the final event computation, although they still contribute to the optimization of the 256-state objective. To guide optimization toward speaker-activity patterns that are truly relevant for event inference, an additional semantic consistency term ($\mathcal{L}_{sem}$) is incorporated into the original loss function:
\vspace{-2mm}
\begin{equation}\label{eq:1}
	\mathcal{L} = \mathcal{L}_{VA} + \mathcal{L}_{VAP} + \mu \cdot \mathcal{L}_{sem} , \quad \mu \in [0.1, 0.2]
\end{equation}
$\mathcal L_{VAP}$ is cross-entropy over 256 VAP microstates, $\mathcal L_{\text{VA}}$ is multi-label binary cross-entropy loss for per-speaker VA outputs, and $\mu$ is an empirically chosen coefficient that controls the new loss contribution. ${L}_{sem}$ is defined as:
\begin{equation}\label{eq:2}
	\mathcal{L}_{\text{sem}} = \mathbb{E}_{t} \left[ -\log \sum_{k \in G_{s_t}} p_t(k) \right], \quad s_t \in \text{\footnotesize $
		\begin{cases} 
			0: \text{spkr1 active} \\
			1: \text{No spkrs active} \\
			2: \text{Both spkrs active} \\
			3: \text{spkr2 active}
		\end{cases} $}
\end{equation}
where $p_t(k)$ is the predicted probability of microstate $k$ at time $t$, $s_t$ the semantic dialogue states and ${G}_{s_t}$ the set of all microstates whose current activity matches the same $s_t$.


Hence, instead of only discriminating among imbalanced and partially redundant microstates, $\mathcal{L}_{sem}$ encourages the model to assign probability mass to all states that share the correct dialogue meaning. From an optimization perspective, this term reduces the influence of rare or ambiguous states, improves training stability, and aligns the learned VAP distribution with the speaker-activity patterns used to derive the target events, leading to more robust predictions throughout training.
\vspace{-1mm}
\begin{table*}[!b]
	\centering
	\footnotesize
	\setlength{\tabcolsep}{3.5pt}
	\renewcommand{\arraystretch}{1.12}
	\setlength{\arrayrulewidth}{0.6pt}
	\begin{tabular}{|c|c|c|c|c|c|}
		\hline
		\multicolumn{1}{|c|}{\multirow{2}{*}{\textbf{Language}}} &
		\multicolumn{1}{c|}{\multirow{2}{*}{\textbf{Backbone}}} &
		\multicolumn{4}{c|}{\rule{0pt}{2.2ex}\textbf{Turn-taking Events (Accuracy / F1)}} \\
		\cline{3-6}
		& & \rule{0pt}{2.2ex}\textbf{S/H} & \rule{0pt}{2.2ex}\textbf{S/L} & \rule{0pt}{2.2ex}\textbf{S-pred} & \rule{0pt}{2.2ex}\textbf{BC-pred} \\
		\noalign{\hrule height 1pt}
		\multirow{2}{*}{English}  & TalkNet & $0.87 \pm 0.03$ / $0.86 \pm 0.01$ & $\mathbf{0.88 \pm 0.01}$ / $0.92 \pm 0.01$ & $0.88 \pm 0.01$ / $0.88 \pm 0.01$ & $0.30 \pm 0.01$ / $0.41 \pm 0.01$ \\
		& WhisperFlamingo & $0.82 \pm 0.08$ / $0.82 \pm 0.09$ & $0.84 \pm 0.02$ / $0.89 \pm 0.02$ & $\mathbf{0.97 \pm 0.04}$ / $0.97 \pm 0.04$ & $0.34 \pm 0.02$ / $0.45 \pm 0.02$ \\
		\hline
		\multirow{2}{*}{German}   & TalkNet & $\mathbf{0.89 \pm 0.01}$ / $0.89 \pm 0.01$ & $0.87 \pm 0.01$ / $0.91 \pm 0.01$ & $0.94 \pm 0.01$ / $0.94 \pm 0.00$ & $0.33 \pm 0.01$ / $0.41 \pm 0.01$ \\
		& WhisperFlamingo & $0.82 \pm 0.04$ / $0.82 \pm 0.04$ & $0.87 \pm 0.01$ / $0.91 \pm 0.01$ & $\mathbf{0.97 \pm 0.02}$ / $0.97 \pm 0.02$ & $0.35 \pm 0.02$ / $0.43 \pm 0.03$ \\
		\hline
		\multirow{2}{*}{French}   & TalkNet & $0.87 \pm 0.03$ / $0.87 \pm 0.03$ & $0.82 \pm 0.03$ / $0.88 \pm 0.02$ & $0.89 \pm 0.05$ / $0.89 \pm 0.05$ & $0.47 \pm 0.03$ / $0.56 \pm 0.03$ \\
		& WhisperFlamingo & $0.84 \pm 0.07$ / $0.84 \pm 0.07$ & $0.85 \pm 0.01$ / $0.89 \pm 0.01$ & $0.96 \pm 0.03$ / $0.96 \pm 0.03$ & $0.42 \pm 0.03$ / $0.52 \pm 0.03$ \\
		\hline
		\multirow{2}{*}{Japanese} & TalkNet & $0.77 \pm 0.01$ / $0.77 \pm 0.01$ & $0.76 \pm 0.01$ / $0.83 \pm 0.01$ & $0.81 \pm 0.01$ / $0.80 \pm 0.01$ & $0.57 \pm 0.01$ / $0.64 \pm 0.01$ \\
		& WhisperFlamingo & $0.86 \pm 0.03$ / $0.87 \pm 0.03$ & $0.72 \pm 0.01$ / $0.80 \pm 0.01$ & $0.86 \pm 0.02$ / $0.86 \pm 0.02$ & $0.56 \pm 0.03$ / $0.62 \pm 0.03$ \\
		\hline
		\multirow{2}{*}{Chinese} & TalkNet & $0.85 \pm 0.01$ / $0.85 \pm 0.01$ & $0.77 \pm 0.01$ / $0.79 \pm 0.01$ & $0.90 \pm 0.01$ / $0.90 \pm 0.01$ & $\mathbf{0.77 \pm 0.01}$ / $0.77 \pm 0.01$ \\
		& WhisperFlamingo & $0.85 \pm 0.01$ / $0.85 \pm 0.01$ & $0.81 \pm 0.04$ / $0.84 \pm 0.04$ & $0.94 \pm 0.01$ / $0.94 \pm 0.01$ & $0.57 \pm 0.01$ / $0.57 \pm 0.01$ \\
		\hline
	\end{tabular}
	\caption{Per-language results on the evaluated corpora. Each cell reports Acc/F1 for the corresponding turn-taking event.}
	\label{tab:per_language_results}
\end{table*}

\vspace{-2mm}
\section{Experiments and results} \label{results}
\subsection{Experimental setup and datasets}
The experimental protocol was based on the original VAP framework, extending it from an acoustic-only setting to a multimodal formulation while preserving the same self-supervised projection objective. The input consists of synchronized 10s context window extracted with a 0.5s slide, prepared according to backbone requirements: audio encoded as Log-Mel spectrograms and the visual stream as detected faces sequence resized to 112×112 grayscale frames. The prediction target is set as 2s future voice activity window, represented through 256 states computed from four temporal bins per speaker (200, 400, 600, and 800 ms). The model predicts future VA from synchronized audio-visual context, and the resulting projections are evaluated against the corresponding zero-shot turn-taking events obtained from the VA annotations, using accuracy and F1-score.

For all the experiments, the data are split at the session level into 80 \% for training, 10 \% for validation, and 10 \% for testing. The datasets were selected for their multilingual coverage and their suitability for the mediator-oriented interaction setting:

\subsubsection{\textbf{NoXi}\protect\linkcite{cafaroNoXiDatabaseMultimodal2017}}

NOvice eXpert Interaction (NoXi) is a multilingual multimodal corpus of screen-mediated dyadic interactions between an expert and a novice. It captures natural knowledge-sharing conversations in a video-conference-like setting, where both participants are recorded in separate rooms with synchronized audio, video, and depth streams. The original corpus contains 84 sessions and around 25 hours of data collected in France, Germany, and the UK, mainly in English, French, and German. NoXi+J \linkcite{funkMultilingualDyadicInteraction2024} extends this setup with 48 Japanese and 18 Chinese sessions under the same protocol, resulting in a total of 150 sessions and about 41 hours of multimodal interaction data.

\subsubsection{\textbf{EDR}\protect\linkcite{cruzWhenHowExpress2025}}
Empathetic Dialogues with a Robot (EDR) consists of semi-structured speaker–listener interactions conducted in the presence of Haru, in which the robot frames each exchange by delivering an introduction and an emotion-based prompt in English, while the humans carry the conversation (see Fig. \ref{fig:scenario}). The dialogues were derived from real-life situations with 79 conversations, 214 exchanges, and 428 labeled utterances.


\subsection{Results}
The following results correspond to experiments designed to evaluate VA-related pretrained backbones adapted with LoRA within their attention mechanisms, in comparison with more generic pretrained backbones and non-pretrained variants presented in Section \ref{related}. All experiments were conducted under the same data distribution and evaluation conditions, with particular emphasis on mediation events relevant to VA projection. The LoRA adapter parameters were kept fixed across experiments to avoid biasing the final outcome comparison. The selection of parameter values was based on their original performance and the proximity to the original tasks for VA projection. In consequence, $\alpha=16$ and $r=16$ for TalkNet and $\alpha=64$ and $r=32$ for WhisperFlamingo, seeking deeper adaptation. Regarding the hyperparameters, the models were trained for $8$ epochs, a linear scheduler for the learning rate with initial value $\mu=0.009$ and the callbacks needed for stable optimization.


Table \ref{tab:per_language_results} summarizes the language-wise results obtained on NoXi+J with the two main pretrained backbones adapted with LoRA. Reported values are represented by the mean $\pm$ standard deviation of accuracy and F1-score over a 4-fold cross-validation. The results show consistent performance across languages, with similar distribution patterns. WhisperFlamingo achieved the highest S-pred scores for English and German, while TalkNet obtained the best S/H score for German and the top BC-pred score for Chinese. In contrast, performance on Japanese remained the lowest overall. Across the remaining languages, TalkNet generally outperformed in S/H and BC-pred metrics, whereas WhisperFlamingo yielded superior results in S/L and S-pred.

Table \ref{tab:multilingual_results} come from models trained with different combinations of NoXi+J languages set.  In addition to the two main pretrained backbones, a third reference model integrating CPC for audio and 3DResNet for video was included as a non-pretrained baseline for comparison, given its well-established strong performance on VA tasks. Across all configurations, the pretrained variants consistently outperformed this reference. WhisperFlamingo tended to report the strongest S-pred values, with similar results for S/H across both backbones.

\vspace{2mm}
\begin{table}[H]
	\centering
	\scriptsize
	\setlength{\tabcolsep}{3.5pt}
	\renewcommand{\arraystretch}{1.28}
	\resizebox{\columnwidth}{!}{%
		\begin{tabular}{|c|c|c|c|c|c|}
			\hline
			\multirow{2}{*}{\textbf{Corpus}} &
			\multirow{2}{*}{\textbf{Backbone}} &
			\multicolumn{4}{c|}{\textbf{Turn-taking Events (F1)}} \\
			\cline{3-6}
			& & \rule{0pt}{2.6ex}\textbf{S/H} & \rule{0pt}{2.6ex}\textbf{S/L} & \rule{0pt}{2.6ex}\textbf{S-pred} & \rule{0pt}{2.6ex}\textbf{BC-pred} \\
			\noalign{\hrule height 1pt}
			\multirow{3}{*}{NoXi}
			& CPC+3DResNet & 0.64 & 0.70 & 0.65 & 0.33 \\
			& TalkNet & 0.94 & 0.93 & 0.91 & 0.42 \\
			& WhisperFlamingo & \textbf{0.97} & 0.91 & \textbf{0.97} & 0.45 \\
			\hline
			\multirow{3}{*}{\begin{tabular}{c}New NoXi+J\\(Chin + Jap)\end{tabular}}
			& CPC+3DResNet & 0.50 & 0.61 & 0.51 & 0.55 \\
			& TalkNet & 0.84 & 0.82 & 0.87 & 0.62 \\
			& WhisperFlamingo & \textbf{0.85} & 0.80 & \textbf{0.93} & 0.66 \\
			\hline
			\multirow{3}{*}{NoXi+J (All)}
			& CPC+3DResNet & 0.55 & 0.59 & 0.58 & 0.39 \\
			& TalkNet & \textbf{0.85} & 0.85 & 0.89 & 0.62 \\
			& WhisperFlamingo & 0.81 & 0.86 & \textbf{0.90} & 0.44 \\
			\hline
			
		\end{tabular}%
	}
	\caption{Experiments on the combined NoXi+J language set using pretrained and baseline backbones.}
	\label{tab:multilingual_results}
\end{table}



Table \ref{tab:comparison_prior_work} compares the F1-scores reported in previous related work on different corpora with those obtained in this study. The first method's results were limited to Shift/Hold events, whereas the other two are directly comparable to the framework evaluated here. In the comparable NoXi+J settings, the results on this work surpassed the reference baselines, particularly for S/H and S-pred, although higher scores in some of the remaining event categories were still reported by the reference methods for specific subsets.

\begin{table}[!h]
	\centering
	\footnotesize
	\setlength{\tabcolsep}{4pt}
	\renewcommand{\arraystretch}{1.05}
	\resizebox{\columnwidth}{!}{%
		\begin{tabular}{lcccccc}
			\multirow{2.5}{*}{\textbf{Method}} &
			\multirow{2.5}{*}{\textbf{Corpus}} &
			\multirow{2.5}{*}{\textbf{Subset}} &
			\multicolumn{4}{c}{\textbf{Turn-taking Events (F1)}} \\
			\cline{4-7}
			\noalign{\vskip 1.7pt}
			& & & \textbf{S/H} & \textbf{S/L} & \textbf{S-pred} & \textbf{BC-pred} \\
			\noalign{\hrule height 0.77pt}
			\textcitelink{russellVisualCuesEnhance2025}{Russell et al.} & Candor & -- & 0.92$^{\dagger}$ & -- & 0.74$^{\ddagger}$ & -- \\
			\hline
			\textcitelink{sagaVoiceActivityProjection2025}{Saga et al.} & NoXi & French & 0.72 & 0.67 & 0.75 & 0.46 \\
			\hdashline[1pt/2pt]
			\multirow{5}{*}{\textcitelink{onishiMultimodalVoiceActivity2025}{Onishi et al.}}
			& NoXi & French   & 0.87 & 0.80 & 0.69 & 0.69 \\
			& NoXi & English  & 0.95 & 0.87 & 0.71 & 0.79 \\
			& NoXi & German   & 0.90 & 0.77 & 0.69 & 0.73 \\
			& NoXi+J & Japanese & 0.82 & 0.74 & 0.65 & 0.48 \\
			& NoXi & All      & 0.90 & 0.81 & 0.72 & 0.71 \\
			\noalign{\hrule height 1.03pt}
			\multicolumn{1}{c}{\multirow{5}{*}{Ours}}
			& NoXi & French   & 0.84 & \lightbold{0.89} & \textbf{0.96} & 0.52 \\
			& NoXi & English  & 0.82 & \lightbold{0.89} & \textbf{0.96} & 0.45 \\
			& NoXi & German   & 0.82 & \lightbold{0.91} & \textbf{0.97} & 0.43 \\
			& NoXi+J & Japanese & \textbf{0.87} & \lightbold{0.80} & \textbf{0.85} & 0.52 \\
			& NoXi & All      & \textbf{0.97} & \lightbold{0.91} & \textbf{0.97} & 0.45 \\
			\hline
		\end{tabular}%
	}
    \caption{Comparison with previous work in terms of downstream turn-taking F1 on the corresponding corpora and subsets. $^{\dagger}$ Keep/Hold and $^{\ddagger}$ Turn/Shift}
	\label{tab:comparison_prior_work}
\end{table}

Table \ref{tab:haru_result} presents experiments on the Haru EDR dataset to validate the proposed approach under the mediation scheme, using the best-performing English configuration for each backbone (Table \ref{tab:per_language_results}). No additional multilingual tests were conducted, as the dataset is English-only and multilingual behavior was already examined in NoXi+J experiments. Because mediation is mainly concerned with floor changes, only Hold, Shift, and S-pred were retained.

\vspace{-0.5mm}
\begin{table}[!h]
	\centering
	\footnotesize
	\setlength{\tabcolsep}{4.5pt}
	\renewcommand{\arraystretch}{1.2}
	\begin{tabular}{lccc}
		\multirow{2}{*}{\textbf{Backbone}} & \multicolumn{3}{c}{\begin{tabular}{c}\rule{0pt}{2.6ex}\textbf{Turn-taking Events}\\\textbf{(Accuracy / F1)}\end{tabular}} \\
		\cline{2-4}
		\noalign{\vskip 1.7pt}
		& \textbf{Hold} & \textbf{Shift} & \textbf{S-pred} \\
		\noalign{\hrule height 0.77pt}
		\noalign{\vskip 1.7pt}
		TalkNet & 0.83 / 0.84 & 0.78 / 0.79 & 0.87 / 0.87 \\
		WhisperFlamingo & 0.92 / 0.92 & 0.83 / 0.85 & 0.91 / 0.91 \\
		\hline
	\end{tabular}
	\caption{Results on the Haru EDR dataset for the mediation-related events.}
	\label{tab:haru_result}
\end{table}
\vspace{-1mm}

\vspace{-0.7mm}
\section{Discussion and future work}
\vspace{0.7mm}
This section examines the contribution of LoRA-adapted VA-related pretrained audio-visual backbones on multimodal VAP for turn-taking prediction.

The results on the NoXi+J dataset presented in Table \ref{tab:comparison_prior_work} showed improvements over the state of the art, especially in S-pred, S/L, and in some subsets of the S/H. These findings demonstrated that specialized backbones can be effectively transferred to the predictive dynamics required for turn-taking. The language-specific results followed a distribution similar to previous work, while reporting stronger performance across all languages. The lower values obtained for Japanese and the highest for English are coherent with the original English training of both backbones and with the language-agnostic trends already reported in the literature. At the same time, and in concordance with previous multilingual findings, the model reported high competitive performance when training jointly on multiple languages together. The F1 values were superior in almost all turn-taking events, which overcome the current state-of-the-art baselines. This was especially representative when combining all available languages in the NoXi dataset. Therefore, although the pretraining language still influences the final outcome, the proposed adaptation strategy demonstrated strong generalization capacity across diverse conversational conditions. A more detailed interpretation of the effects of language on the model is beyond the scope of this study.

Another relevant aspect was the contrast between pretrained and non-pretrained backbone influence. Even though many intermediate experiments were omitted from this study due to the large number of possible combinations, it is important to note that the selected pretrained backbones surpassed state-of-the-art results after the LoRA strategy was implemented. In fact, this tendency could already be inferred from the lower results that the frozen version of the original CPC + 3DResNet provided and justified the need of an adaptation of the original VA-related task to the MM-VAP. From an experimental perspective, all reported results in this work were obtained under the same conditions to avoid favoring any setting with its own best configuration and to ensure a fair and rigorous comparison across inner variants and with previous work. However, it is worth noting that additional tests suggested that further gains may still be achieved when the hyperparameters are adjusted to particular backbone origins and language conditions, which may be relevant for final real-world applications.

Regarding the analysis of the results obtained on the Haru EDR dataset, some event analyses were omitted due to the lack of useful information for mediation, where only turn-taking events related to floor change and floor development matter. In this context, the results also provided positive validation of the proposed approach for capturing the conversational dynamics most relevant to the target HRI scenario. This supports the feasibility of the MM-VAP model for turn-taking management in a mediator-oriented robotic role, where the robot is expected to monitor and anticipate the evolution of conversational floor among at least two interlocutors. Nonetheless, complete validation of this approach should go beyond the current offline methodology. Consequently, the main limitation of the present study is that, as in previous work, the proposed approach was not validated under real-time MM-VAP conditions. This remains challenging because MM-VAP requires high computation resources, especially when large pretrained encoders must operate together with other high-demanding modules. In addition, the effective integration of the anticipatory capabilities of MM-VAP into the multimodal dialogue system introduces further complexity. To move toward this objective, future work should prioritize lightweight backbones that better satisfy real-time constraints while preserving similar performance achieved offline. A further promising research direction would be to extend the current audio-visual formulation into a complete trimodal architecture that incorporates contextual information, given its relevance in turn-taking behavior.

Overall, the present findings indicated that the combination of specialized pretrained audio-visual encoders, together with their original attention capabilities and LoRA adaptation, provides an effective framework for multimodal VAP. 

\vspace{0.5mm}
\section{Conclusions}
\vspace{-1mm}

For social robots, natural turn-taking modeling is more appropriately addressed by anticipating the speaker's future activity rather than reacting only to pauses or end-of-turn silences. Nowadays, VAP represents the most promising framework, with additional value when extended to the multimodal domain to compensate for limitations that audio-only solutions still present in addressee identification and floor-transition anticipation. In this direction, this work has presented a multimodal voice activity projection model for dyadic interaction, motivated by the Haru mediation requirements. The proposed transformer-based architecture relied on specialized backbones that were adapted more efficiently to the VA forecasting task using LoRA. The obtained results indicated that transferring VA-related audio-visual backbones within its original learned inter-modality attention capabilities is more efficient than using generic encoders or hand-crafted feature extensions. This was supported by the model surpassing the state of the art in several turn-taking events, especially those linked to mediation, as well as by its validation on the available Haru EDR dataset. Furthermore, the solution exhibited robustness across languages and remained consistent under multilingual training, achieving higher values than previous work.

Although the current evaluation was restricted to recorded conversational data, the results indicate that this approach is suitable for conversational floor management in HRI, particularly in settings that require anticipating speaker transitions and identifying appropriate moments for intervention. Therefore, the proposed pipeline suggests strong potential for this direction to improve social robot behavior involving human-human communication.


\vspace{3mm}
\noindent\textbf{Acknowledgments:} This work was partially funded by the Spanish Ministry of Science and Innovation and the State Research Agency (MCIN/AEI/10.13039/501100011033) under the projects \textbf{TIFON} [MIG-20232039 / PLEC2023-010251] (\hyperref[author:AC]{A.C.}, \hyperref[author:GP]{G.P.}), \textbf{PICRAH4.0} [PLEC2023-010353] (\hyperref[author:LM]{L.M.}) and \textbf{LIPTACON}, by CDTI Innovation through the \textit{Programa Tecnológico Espacial (PTE) 2024} via project [PTEP-20241001] (\hyperref[author:AC]{A.C.}, \hyperref[author:GP]{G.P.}). 



\vspace{4mm}
{\setlength{\parskip}{0pt}
	\bibliographystyle{ieeetr}
	\bibliography{references}
}

\end{document}